# When do drivers concentrate? Attention-based driver behavior modeling with deep reinforcement learning

Xingbo Fu[†]
System Engineering Institute
Xi'an Jiaotong University
Xi'an, China
xbfu1994@gmail.com

Feng Gao
System Engineering Institute
Xi'an Jiaotong University
Xi'an, China
fgao@sei.xjtu.edu.cn

Jiang Wu
System Engineering Institute
Xi'an Jiaotong University
Xi'an, China
jwu@sei.xjtu.edu.cn

## ABSTRACT

Driver distraction a significant risk to driving safety. Apart from spatial domain, research on temporal inattention is also necessary. This paper aims to figure out the pattern of drivers' temporal attention allocation. In this paper, we propose an actor-critic method – Attention-based Twin Delayed Deep Deterministic policy gradient (ATD3) algorithm to approximate a driver's action according to observations and measure the driver's attention allocation for consecutive time steps in car-following model. Considering reaction time, we construct the attention mechanism in the actor network to capture temporal dependencies of consecutive observations. In the critic network, we employ Twin Delayed Deep Deterministic policy gradient algorithm (TD3) to address overestimated value estimates persisting in the actor-critic algorithm. We conduct experiments on real-world vehicle trajectory datasets and show that the accuracy of our proposed approach outperforms seven baseline algorithms. Moreover, the results reveal that the attention of the drivers in smooth vehicles is uniformly distributed in previous observations while they keep their attention to recent observations when sudden decreases of relative speeds occur. This study is the first contribution to drivers' temporal attention and provides scientific support for safety measures in transportation systems from the perspective of data mining.

## 1 INTRODUCTION

Driver inattention is a major factor in many car and truck crashes and incidents. Driver distraction happens because a driver is temporarily focusing on an object, task or event not related to driving, which reduces the driver's awareness and decision making ability [1]. Several studies have demonstrated that external factors such as roadside advertising signs affect drivers' visual attention [2, 3]. To detect a driver's attention, driver distraction detection systems employ driver cameras and sensors monitoring a driver's head pose and eye movements [4, 5, 6]. Nevertheless, such explanations about a driver's attention allocation only focus on spatial domain and these systems work only after driver distraction happens. Apart from instantaneous spatial factors, a driver's previous observations also influence the driver's decision at current time step in the real world.

This fact has attracted attention of several researchers attempting to capture driver behaviors in modeling car-following behavior. Some researchers proposed data-driven methods to learn driver behaviors from empirical car-following data. The recurrent neural network (RNN) was employed to capture temporal dependencies for sequential historical observations as inputs [7]. Furthermore, Long Short Term Memory (LSTM) neural network and its variation Gated Recurrent Unit (GRU) neural network have been widely applied since they outperform feedforward neural network (FNN) and RNN [8]. However, these driver behavior simulation approaches fail to construct long-term correlations for the whole vehicle trajectory in car-following model.

In deep reinforcement learning (RL), an agent obtains more accumulated rewards by interacting with the environment. Considering car-following model as continuous control, actor-critic methods with deep deterministic policy gradient (DDPG) take the driver's previous observations as inputs of the actor network and it outputs longitudinal acceleration of the vehicle [9]. One major drawback of this approach is that temporal dependencies are not completely extracted through FNN; therefore, we still have no conclusion about the driver's attention distribution in consecutive time steps. In addition, another problem with this approach is overestimated value estimates in DDPG [10].

Attention mechanism [11] is an influential idea in deep learning and it has been applied in various problems such as image processing and neural machine translation. Attention mechanism calculates the weight of each feature and less useful features should be assigned a lower weight as they contribute less to the output [12]. Hence, attention mechanism can figure out a driver's attention allocation for consecutive time steps.

To tackle all the aforementioned issues and challenges, we propose an actor-critic structure – Attention-based Twin Delayed Deep Deterministic policy gradient (ATD3) algorithm in this paper. The introduction of attention mechanism can capture temporal dependencies among consecutive observations and TD3 overcomes overestimated value estimates in DDPG.

A case study is conducted to show that the ATD3 algorithm outperforms the state-of-the-art approaches in modeling car-following behavior. We evaluate the interpretability of the weight vector through qualitative analysis. Finally, we present effective approaches about deceleration zone and roadside advertising to improve driving safety.

Compared to previously published work on car-following behavior and a driver's attention, main contributions of this paper are:
- We propose the ATD3 algorithm which can accurately model car-following behavior based on historical car-following data;
- We show that the ATD3 algorithm outperforms existing methods in modeling car-following behavior on Next Generation Simulation (NGSIM) Vehicle Trajectories and Supporting Data;
- The results in this paper indicates that a driver's attention mainly distributes in the recent 0.8 second and is sensitive to abrupt decreases of the relative speed between the driver own and the lead vehicle (LV);
- This paper provides scientific support for safety measures in transportation systems from the perspective of data mining. According to the results, we present several approaches which can reduce driver inattention.

The rest of this paper is organized as follows: In Section 2, we introduce the background of modeling car-following behavior. Section 3 presents the details of the proposed ATD3. The experiments on real world are presented in Section 4 and Section 5 contains the results and discussion. Section 6 concludes the paper.

## 2 BACKGROUND

### 2.1 Car-following model

A car-following model describes the driver behavior of a following vehicle (FV) according to an observation the driver detects in the FV, such as gaps and relevant speeds between the FV and the LV. Based on each state, the FV performances a continuous action and results in a subsequent observation. The car-following model aims to simulate driver behavior in dynamic situations and minimize the disparity between the values of simulated and observed behavior.

In car-following model, the observation at a certain time step $t$ includes three key parameters: the absolute speed of FV $V_F(t)$, relative speed $\Delta V(t) = V_L(t) - V_F(t)$ and gap $\Delta S(t) = S_L(t) - S_F(t)$ between the LV and the FV. The action is the longitudinal acceleration of the FV $a(t)$. With the observation and action at time step $t$, a kinematic point-mass model was used for observation updating as follows

$$V_F(t+1) = V_F(t) + a(t) \cdot \Delta T$$
$$\Delta V(t+1) = V_L(t+1) - V_F(t+1) \qquad (1)$$
$$\Delta S(t+1) = \Delta S(t) + \frac{\Delta V(t) + \Delta V(t+1)}{2} \cdot \Delta T$$

where $\Delta T$ is the simulation time interval and is set as 0.1 second, and $V_L(t)$ is the velocity of LV at time $t$, which was the externally input.

The task of this paper is to accurately modeling car-following behavior and fathom a driver's attention allocation in consecutive time steps.

### 2.2 RL-based car-following model

In RL, an agent generates the longitudinal acceleration $a(t)$ according to the state which is determined by the FV and the LV. Then the FV executes the acceleration $a(t)$ following the kinematic point-mass model as shown in Equation (1) and therefore we can get the new state. In order to model this car-following behavior accurately, this agent is trained based on data from a replay buffer. The whole process can be shown in Figure 1.

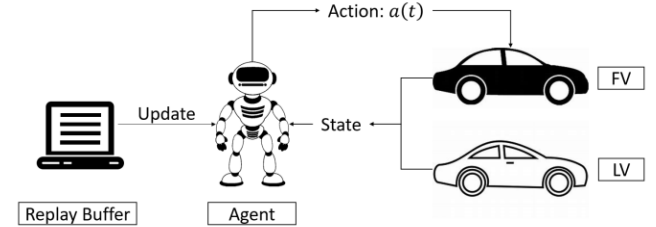

**Figure 1: RL-based car-following model**

## 3 METHODOLOGY

Figure 2 shows the structure of our approach ATD3. As shown in Figure 2, ATD3 contains a single actor network and a pair of critic networks. The input of the actor network is the state $s_t$ at time step $t$ and it outputs an action $a_t$ through an attention mechanism. The state $s_t$ at time step $t$ consists of $T$ observations $(obs_{t-T+1}, \cdots, obs_{t-1}, obs_t)$ from time step $t-T+1$ to $t$. The observation at time step $t$ includes three key parameters: the absolute speed of FV $V_F(t)$, relative speed $\Delta V(t)$ and gap $\Delta S(t)$ between the LV and the FV

$$obs_t = (V_F(t), \Delta V(t), \Delta S(t)) \qquad (2)$$

In the pair of critic networks, we get two separate outputs $Q_{\theta_1}$ and $Q_{\theta_1}$ with respect to the state $s_t$ as well as the action $a_t$ generated by the actor network.

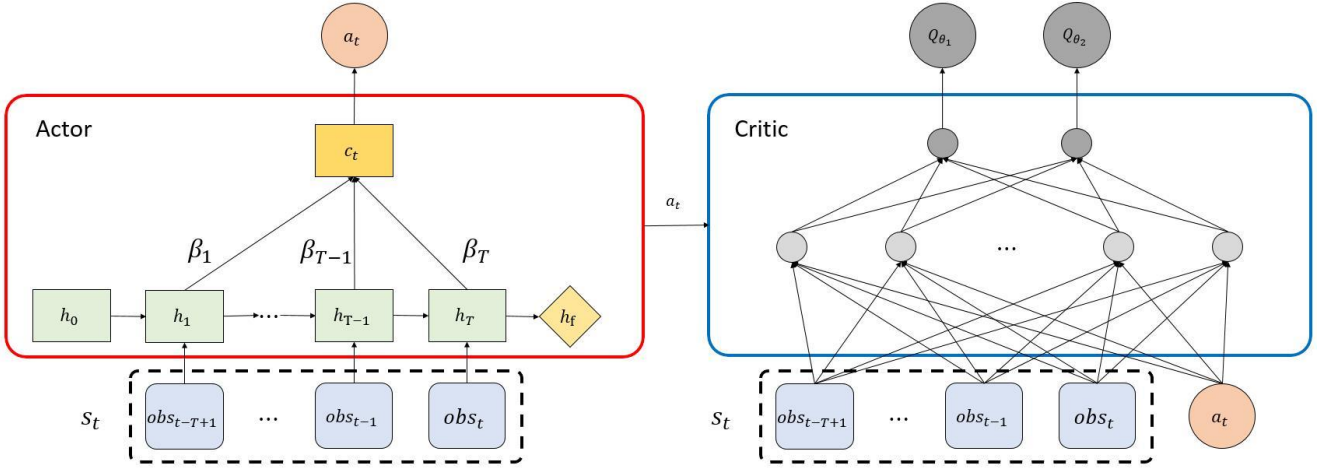

**Figure 2: Structure of ATD3**

## 3.2 TD3

In RL, an agent interacts with the environment and learns an optimal policy for sequential decision making problems. In the problem with continuous action spaces, the deep deterministic policy gradient (DDPG) algorithm uses a learned value estimate to generate a continuous action according to the state [13].

DDPG is an approach based on Actor-Critic algorithm and generally consists of two separate networks: an actor network and a critic network. Based on its input – the current state $s_t$, the actor network with weights $\varphi$ outputs an action $a_t$ with respect to its policy $\pi_\varphi(s_t)$. Therefore, the agent gets a reward $r_t$ for the action choice and moves to the next state $s_{t+1}$ according to the environment. Taking the action $a_t$ generated by the actor network and the current state $s_t$, the critic network with weights $\theta$ outputs the expected return aiming to approximate the value function. In Q-learning, the value function is learned based on Bellman equation

$$Q^\pi(s_t, a_t) = r + \gamma E_{s_{t+1}, a_{t+1}}[Q^\pi(s_{t+1}, a_{t+1})] \quad (3)$$

where

$$a_{t+1} = \pi(s_{t+1}) \quad (4)$$

According to Bellman equation, we consider minimizing the loss function $L(\theta)$ parameterized by $\theta$

$$L(\theta) = \frac{1}{N}\sum(y_t - Q_\theta(s_t, a_t))^2 \quad (5)$$

where

$$y_t = r_t + \gamma Q_\theta(s_{t+1}, a_{t+1}), \ a_{t+1} = \pi_\varphi(s_{t+1}) \quad (6)$$

Different from DDPG, TD3 maintains a single actor network and a pair of critic networks with weights $\theta_1$ and $\theta_2$ to eliminate overestimation bias in DDPG. We update the pair of critic networks towards the minimum target value of actions with respect to the target policy $\pi_{\varphi'}(s_{t+1})$

$$y_t = r_t + \gamma \min_{k=1,2} Q_{\theta_k}(s_{t+1}, a_{t+1} + \epsilon) \quad (7)$$

where

$$a_{t+1} = \pi_{\varphi'}(s_{t+1}) \quad (8)$$

and $\epsilon$ is a small amount of random noise.

Hence, we update the critic network of TD3 through minimizing the loss function $L(\theta)$ where $y_t$ can be calculated by Equation 5.

The actor network can be updated through deterministic policy gradient algorithm

$$\nabla J(\varphi) = \frac{1}{N}\sum \nabla_{a_t} Q_{\theta_1}(s_t, a_t)|_{a_t=\pi_\varphi(s_t)} \nabla_\varphi \pi_\varphi(s_t) \quad (9)$$

## 3.3 Attention mechanism in TD3

In some RL-based car-following behavior models, actor network chooses an action $a_t$ with the current state $s_t$ at timestamp $t$ as the input. In the real world, however, drivers consider the current state as well as several former states when they decide to brake or accelerate. Hence, considering reaction time steps $T$ is indispensable for actor network to simulate driver behavior.

In order to figure out a driver's attention allocation for consecutive time steps, we employ attention mechanism as actor network to approximate the policy instead of using fully-connected network.

As shown in Figure 1, the hidden state $h_t$ in the encoder is computed as follows

$$h_t = \varphi(obs_{t-1} U^E + h_{t-1} W^E) \quad (10)$$

where $U^E$ and $W^E$ are the learnable weight matrices in the encoder. $\varphi$ is a tanh activation. $h_0$ is the zero state.

The context vector $c_t$ is the dynamically weighted average over all the hidden states of the encoder calculated by attention mechanism as follows

$$c_t = \sum_{j=t-T+1}^{t} \beta_j h_j \quad (11)$$

where $\beta_j$ is an element of weight vector $\beta \in \mathbb{R}^{1 \times T}$ and is computed as follows

$$\beta_j = \frac{exp(score(h_f, h_j))}{\sum_{j=t-T+1}^{t} exp(score(h_f, h_j))} \quad (12)$$

where $h_f$ is the final hidden state of the encoder.

As Equation (12) shows, $\beta_j$ is a softmax result of a score function. According to Luong attention mechanism, the score function has three different alternatives. In this paper, we implement the concatenation-based attention [14] mechanism as follows

$$score(h_f, h_j) = W_2^a \tanh(W_1^a[h_f; h_j]) \quad (13)$$

Finally, we can get the action:

$$a_t = \varphi(c_t W^c) \quad (14)$$

where $W^c$ is the learnable weight matrix in the actor network and $\varphi$ is a tanh activation.

### 3.4 Experience replay and target network

In order to address the issue of independently and identically distributed among training samples in RL, we employ a replay buffer with fixed size in the ATD3 algorithm [13]. The replay buffer is a circular queue with fixed length and stores transitions generated with respect to the actor network and Equation 1. During training, we randomly sample transitions $(s_i, a_i, r_i, s_{i+1})$ from the replay buffer. If the replay buffer is full, the oldest transitions will be replaced.

To avoid divergence between the predicted value and target value, we use a separate network called a target network for calculating the target value [13]. In the ATD3 algorithm, we deploy three target networks $\theta_1'$, $\theta_2'$ and $\varphi'$ for the main actor network and critic networks, respectively. The target networks are initialized in the same structure as the main actor network and critic networks. The target networks' weights are updated by small amounts towards the main network (soft target updates); therefore, the ATD3 algorithm is more stable.

### 3.5 ATD3

Therefore, we train the ATD3 algorithm for modeling car-following behavior as shown in Algorithm 1.

## 4 EXPERIMENTS

This section presents the implement of our ATD3 algorithm as well as seven baseline algorithms in the real-world experiments. And we open source our code and results on https://github.com/xbfu/ATD3.

### 4.1 Dataset

The NGSIM program collected detailed vehicle trajectory data on southbound US 101 and Lankershim Boulevard in Los Angeles, CA, eastbound I-80 in Emeryville, CA and Peachtree Street in Atlanta, Georgia [15]. Data was collected through a network of synchronized digital video cameras. This vehicle trajectory data contains the precise location of each vehicle within the study area every 0.1 second, resulting in detailed lane positions and locations relative to other vehicles.

In order to model car-following behavior properly, the vehicle trajectory used in this paper need to satisfy the following criteria [16]:

- The identification of the LV remains constant to ensure the FV follows the same LV in a certain trajectory;
- The distance between the FV and the LV is less than 120 meters in each trajectory to eliminate free-flow traffic scenarios;
- The lateral distance between the FV and the LV is less than 2.5 meters in each trajectory to guarantee that they drive in the same lane;
- Each trajectory lasts longer than 15 seconds to ensure the trajectory is long enough to be analyzed.

According to the criteria, we select the data of 600 different vehicles and each trajectory lasts for 400 time steps. The training data consists of records of 450 vehicles from these 600 vehicles (180,000 time steps in all) and the rest are test data. The observations of a driver (speed of FV, relative speed and gap between the driver and the LV) for 400 time steps are shown in Figure 3.

---

**Algorithm 1**: ATD3 for modeling car-driver behavior

Initialize critic networks $Q_{\theta_1}$, $Q_{\theta_1}$ and actor network $\pi_\varphi$ with random parameters $\theta_1$, $\theta_2$, $\varphi$
Initialize target networks $\theta_1' \leftarrow \theta_1$, $\theta_2' \leftarrow \theta_2$, $\varphi' \leftarrow \varphi$
Set up empty replay buffer B
Initialize state $s_1 = (obs_1, \cdots, obs_9, obs_{10})$
**for** t = 1 **to** N
  Select action with exploration noise
$$a_t \sim \pi_\varphi(s_t) + \epsilon, \quad \epsilon \sim N(0, \sigma)$$
  Calculate reward $r_t$ and new observation $obs_{new}$
  New state $s_{t+1} = [s_t(2:10); obs_{new}]$
  Store transition $(s_t, a_t, r_t, s_{t+1})$ into replay buffer B
  Sample minibatch of $m$ transitions $(s_i, a_i, r_i, s_{i+1})$ from B
$$a_{i+1} = \pi_{\varphi'}(s_{i+1})$$
  Set $y_i = r_i + \gamma \min_{k=1,2} Q_{\theta_k}(s_{i+1}, a_{i+1} + \epsilon)$
  Update critics $\theta_i \leftarrow \arg\min_{\theta_i} \frac{1}{N} \sum (y_i - Q_{\theta_i}(s_i, a_i))^2$
  **if** t mod 2 **then**
    Update actor policy using deterministic policy gradient
$$\nabla J(\varphi) = \frac{1}{N} \sum \nabla_{a_i} Q_{\theta_1}(s_i, a_i)|_{a_i = \pi_\varphi(s_i)} \nabla_\varphi \pi_\varphi(s_i)$$
    Update target networks
$$\theta_1' \leftarrow \tau\theta_1 + (1-\tau)\theta_1'$$
$$\theta_2' \leftarrow \tau\theta_2 + (1-\tau)\theta_2'$$
$$\varphi' \leftarrow \tau\varphi + (1-\tau)\varphi'$$
  **end if**
**end for**

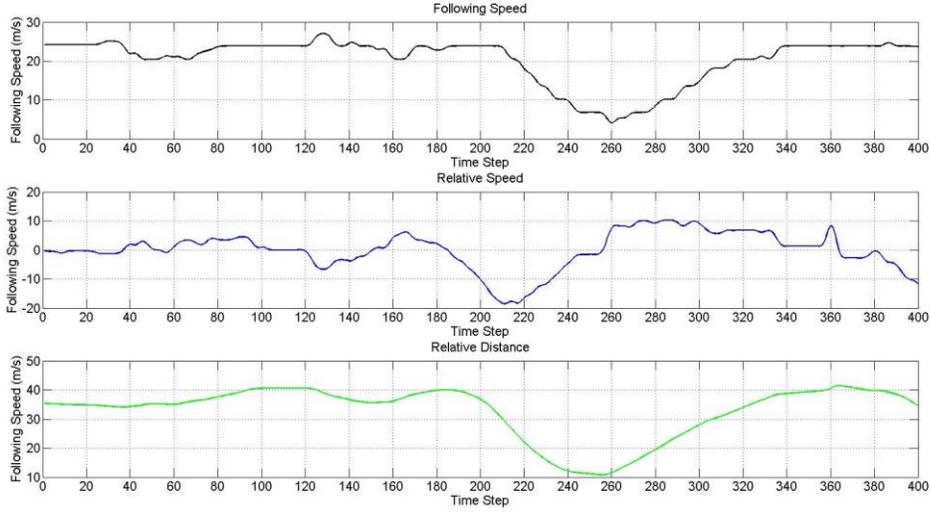

**Figure 3: Observations of a driver at each time step**

## 4.2 Experiment Settings

The learning rates of actor and critic are set to $10^{-3}$ and $10^{-3}$ respectively. The discount factor is 0.99 and the update rate of target networks is $10^{-3}$. During the exploration, the zero-mean Gaussian noise with variance 0.1 is added to the action. The training process contains 60 epochs and batch size is fixed to 200; therefore, each epoch has 60 cycles. The agent generates 200 samples and then these samples are stored in the experience replay buffer with length of $10^5$. The momentum-based algorithm called Adam is used to optimize the loss function in the training. Reaction time is fixed to 10 time steps (1 second). In ATD3, the hidden sizes of actor and critic are both 100. Similarly, the hidden sizes of all the neural network-based baseline algorithms are also 100. In car-following model, the reward function is defined as

$$r_t = \log\left(\left|\frac{V^{sim}(t)-V^{obs}(t)}{V^{obs}(t)}\right|\right) \quad (15)$$

where $V^{sim}(t)$ is the simulated speed and $V^{obs}(t)$ is the observed speed.

To measure the effectiveness of our proposed model and baseline algorithms, we use the root mean square percentage error (RMSPE) of speed was adopted as the evaluation metric

$$\text{RMSPE} = \sqrt{\frac{\sum_{i=1}^{N}\left(V^{sim}(t)-V^{obs}(t)\right)^2}{\sum_{i=1}^{N}\left(V^{obs}(t)\right)^2}} \quad (16)$$

## 4.2 Baseline algorithms

We compare our proposed model with the following seven baseline algorithms.

- IDM [17]: As a typical desired measures model, the intelligent driver model (IDM) considers both the desired speed and the desired following distance. In this experiment, a genetic algorithm (GA) is used to calibrate the IDM in this work.
- ANN: We employ a fully-connected network with one hidden layer to approximate the action according to the current state.
- ANNRT: In this experiment, considering reaction time, ANNRT is also implemented and its inputs contain the states of ten time steps.
- GRU [18]: A gated recurrent unit (GRU) is part of a specific model of recurrent neural networks which use recurrent structure to replace the fully-connected network in ANN so that GRU can effectively capture temporal dependencies among ten time steps.
- Attn [11]: Attention mechanism is added to RNN and the context vector replaces the hidden states in RNN.
- DDPG [13]: The basic DDPG with current states as inputs.
- DDPGRT [9]: This method inputs the states of ten time steps with respect to reaction time.

## 5 RESULTS AND DISCUSSION

This section analyzes results of the experiments in this paper. The results contain performance of ATD3 and the other baseline algorithms as well as the driver's attention allocation for consecutive time steps.

### 5.1 Performance on RMSPE

Table 1 shows RMSPE of ATD3 and the other baseline algorithms on NGSIM dataset for the car-following driver behavior models.

**Table 1: RMSPE of Car-following Models**

| NO. | Method | RMSPE (%) |
|---|---|---|
| 1 | IDM | 13.28 |
| 2 | ANN | 9.04 |
| 3 | ANNRT | 8.56 |
| 4 | RNN | 8.50 |
| 5 | Attn | 8.09 |
| 6 | DDPG | 8.10 |
| 7 | DDPGRT | 7.82 |
| 8 | ATD3 | 7.55 |

In Table 1, we can observe that ATD3 outperforms the other seven baseline approaches. The traditional car-following model – IDM is the worst among these algorithms and RL-based algorithms are better than neural network-based algorithms. The attention mechanism improves the performance of car-following model.

### 5.2 Performance on Stable Following

A proper Car-following model is supposed to be stable and safe. Figure 4 shows the simulated speed of driver 1 of car-following models. The ATD3 algorithm is stabler and safer. The simulated speed of DDPGRT after time step 330 gets faster than the real-world speed, which leads to unsafe distance between the driver and the LV.

### 5.3 Driver Attention Allocation

One significant purpose of this study is to detect when drivers concentrate when they are in following cars. We record the weight vector $\beta$ for each time step of and the results of driver 2 are shown in Figure 5.

Figure 5 illustrates that a driver's attention mainly distributes in the last 8 time steps (0.8 second). In addition, a driver's attention allocation fluctuates in different time steps. Specifically, the driver's attention concentrates on the latest two or three time steps at the time step 85 and 370.

In order to figure out the scenario when a driver's attention centers on the latest few time steps, Figure 6 shows driver 2's observations (speed of the following car, relative speed and relative distance) at each time step.

As shown in Figure 6, a driver's attention is sensitive to a sudden decrease of relative speed. This decrease occurs when the driver in the lead car abruptly brakes.

### 5.3 Discussion

According to the results of a driver's attention allocation, we draw these three conclusions:

- Recent 8 observations (0.8 second) keep drivers' main attention. The settings with longer reaction time may impair the ability of neural networks.
- A sudden decrease of relative speed always holds a driver's most attention. When it occurs, drivers focus on recent 2 or 3 observations regardless of early observations.
- When a lead vehicle and a following vehicle both have smooth rides, the attention of the driver in the following vehicle is almost uniformly distributed in observations.

With respect to the three conclusions, we can keep drivers concentrating in the following ways:

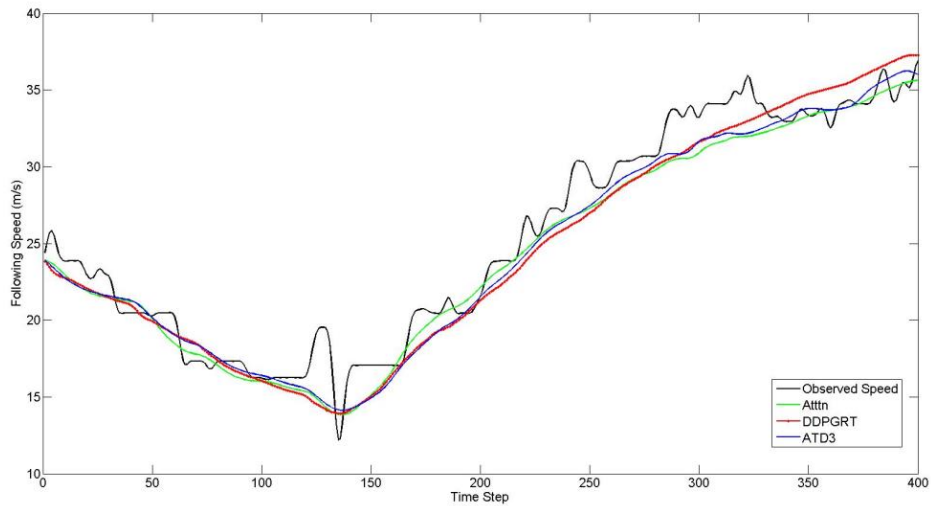

**Figure 4: Simulated speed of car-following models**

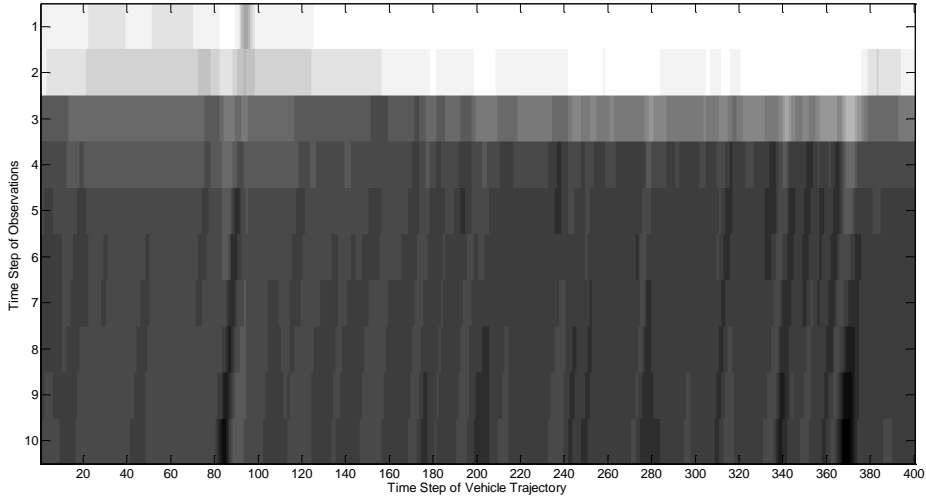

**Figure 5: Weight vector at each time step**

- **Deceleration zone.** Smooth rides result in the uniform driver attention allocation and drivers fail to focus on the current observations. Considering a driver's attention sensitivity about a sudden brake of the lead car, employing some deceleration zone is necessary. According to the results in this paper, effective deceleration zones are employed 0.3-second-drive distance before those locations where traffic accidents frequently occur.
- **Roadside advertising.** Different from deceleration zone, we should reduce roadside advertising situated along straight roads. Smooth rides on straight roads leads to driver distraction from the driver's current observations and roadside advertising will exacerbate this problem.
- **Roadside advertising with deceleration zone.** Roadside advertising may distract drivers. Before drivers can clearly recognize roadside advertising, we can set deceleration zone to draw a driver's attention back to the current state in case they are distracted by both spatial factors and previous observations.

## 6 CONCLUSION

In this paper, we propose the ATD3 algorithm – an attention-based actor-critic method for modeling car-following behavior. The ATD3 algorithm outperforms the state-of-the-art baseline algorithms in the real-world experiments. In addition, this paper fathoms the driver attention allocation for consecutive time steps with the ATD3 algorithm. A driver's attention mainly distributes in the recent 0.8 second and is sensitive to abrupt decreases of the relative speed between the driver own and the LV. Finally, three pieces of advice about deceleration zone and roadside advertising are given with respect to these discoveries. This paper is the first research of driver attention allocation in temporal time steps and provides scientific support for theories in transportation systems.

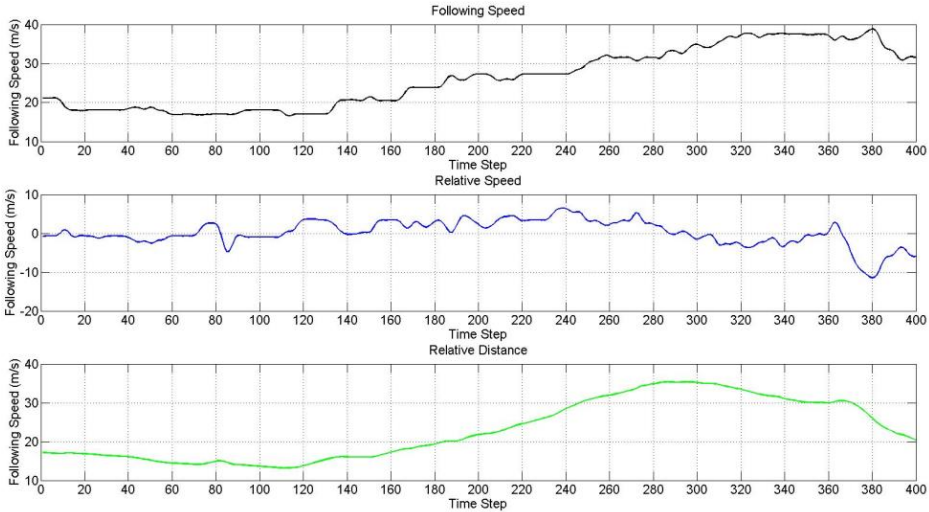

**Figure 6: Observations at each time step**